\documentclass[conference]{IEEEtran}
\IEEEoverridecommandlockouts
\usepackage{cite}
\usepackage{amsmath,amssymb,amsfonts}
\usepackage{algorithmic}
\usepackage{graphicx}
\usepackage{subcaption} 
\usepackage{hyperref}   
\usepackage{textcomp}
\usepackage{xcolor}
\def\BibTeX{{\rm B\kern-.05em{\sc i\kern-.025em b}\kern-.08em
    T\kern-.1667em\lower.7ex\hbox{E}\kern-.125emX}}
\begin{document}

\title{PID Controller Optimization for Low-cost Line Follower Robots}

\author{\IEEEauthorblockN{Samet Oguten\IEEEauthorrefmark{1}, Bilal Kabas\IEEEauthorrefmark{1}}
\IEEEauthorblockA{\IEEEauthorrefmark{1}Department of Electrical-Electronics Engineering,
Abdullah Gül University\\
Kayseri, Turkey\\}
Email: samet.oguten@agu.edu.tr,
bilal.kabas@agu.edu.tr
}

\maketitle

\begin{abstract}
In this paper, modification of the classical PID controller and development of open-loop control mechanisms to improve stability and robustness of a differential wheeled robot are discussed. To deploy the algorithm, a test platform has been constructed using low-cost and off-the-shelf components including a microcontroller, reflectance sensor, and motor driver. This paper describes the heuristic approach used in the identification of the system specifications as well as the optimization of the controller. The PID controller is analyzed in detail and the effect of each term is explained in the context of stability. Lastly, the challenges encountered during the development of controller and robot are discussed. Code is available at: \emph{\href{https://github.com/sametoguten/STM32-Line-Follower-with-PID}{https://github.com/sametoguten/STM32-Line-Follower-with-PID}}.
\end{abstract}


\section{Introduction}

Navigation is an essential part of Robotics applications. To complete a specific task, it is crucial for a robot to follow a specific path. In advanced applications, robots are equipped with sophisticated sensors such as depth cameras, radars and lidars for enhanced perception capabilities. Availability of such sensors enables the development of more sophisticated and high-fidelity localization and mapping algorithms. However, these sensors are expensive. Some certain tasks requiring repetitive motion in a relatively fixed environment can be done by cost-effective applications. For instance, a line tracking robot having only an array of reflectance sensors \cite{Balaji2015}, \cite{Latif2020}, \cite{Hasan2012}, \cite{Pakdaman2010} can be used to carry a specific load from one point to another in a factory. The aim of this research is to develop a controller algorithm for a differential wheeled robot to follow a specific path. There are two main requirements. The first requirement is that the robot should complete the track. The second one is that the completion time should be as short as possible. Throughout this research, only off-the-shelf and cost-efficient components have been used.

PID control stand for proportional, integral, and derivative. PID controller is a closed-loop control system widely used in industry. It tries to bring the system to the state specified by the input by calculating the error. Error calculation is done by comparing the reference input with the output of the system. If the proportional gain in the PID control is increased, the response of the system to disturbances becomes faster. However, after a certain level, increasing the proportional gain can cause the system to show an oscillatory unstable behavior. The integral control in the PID controller is used to keep track of the past errors in the system. Thus, steady-state error is eliminated. However, increasing integral gain can cause overshooting. The resulting overshoot can cause the system to show an unstable behavior. The main effect of the derivative controller is that it minimizes overshooting and speeds up the transient response. It can be thought of as a mechanism to detect future errors in the system. As the integral control minimizes the steady-state error at the cost of overshooting, derivative control can be used to minimize that overshooting to increase the robustness of the system. However, a derivative controller may cause the system to be unstable in the presence of noise as differentiation amplifies high frequency signals.

In this research, PID controller has been used to develop a high-speed and high-fidelity line tracking controller. The PID controller calculates errors according to the data coming from the reflectance sensor array. The main reason why PID control is used during the development of the line tracking controller is to prevent overshooting, minimize steady-state error, and oscillation that may occur in the system \cite{Franklin1994}.

\section{Test Platform}

\subsection{Mechanical Components}

\begin{figure*}[t]
\centering
\subcaptionbox{Top view}{\includegraphics[width=0.32\linewidth]{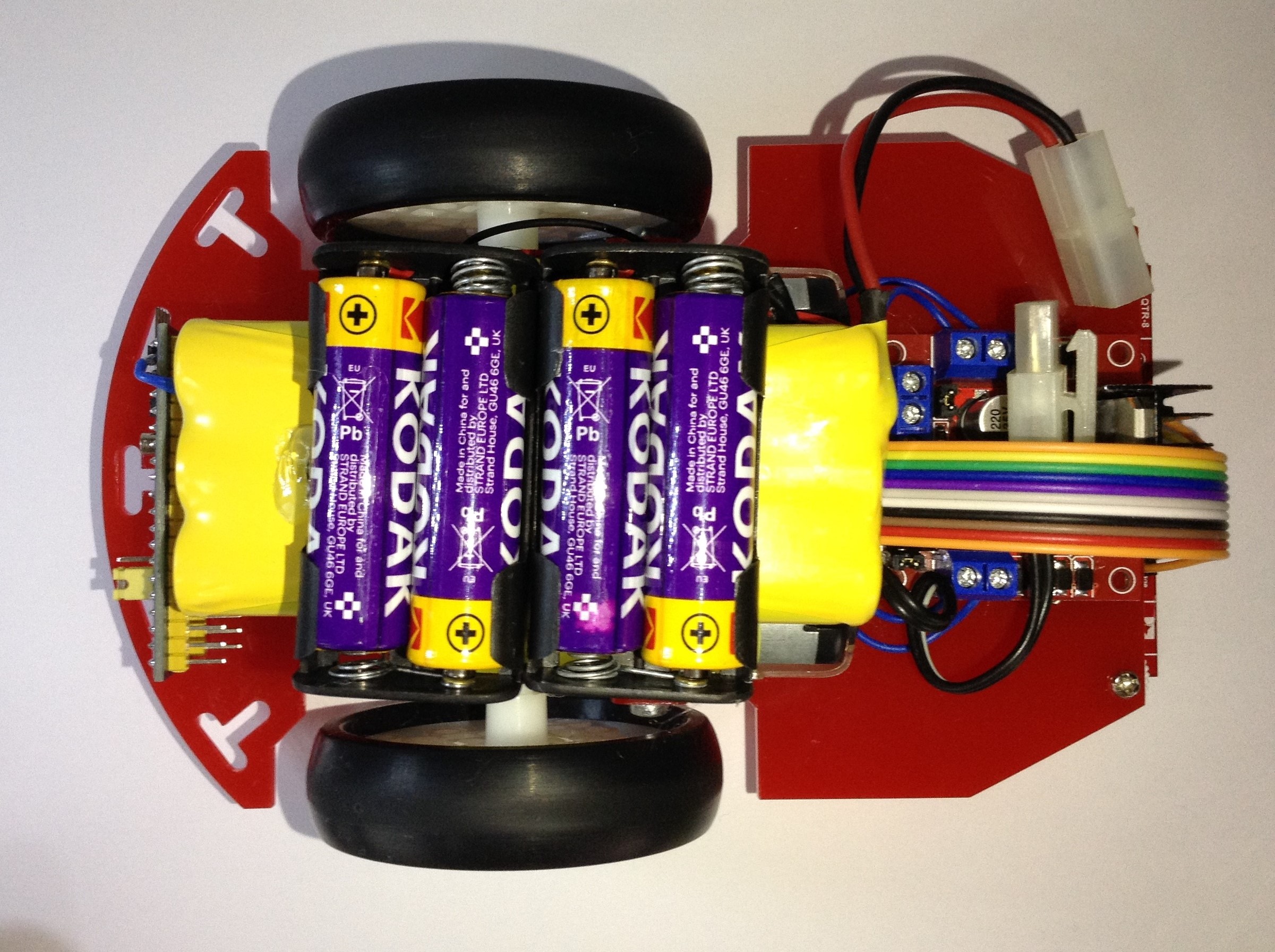}}%
\hfill
\subcaptionbox{Side view}{\includegraphics[width=0.32\linewidth]{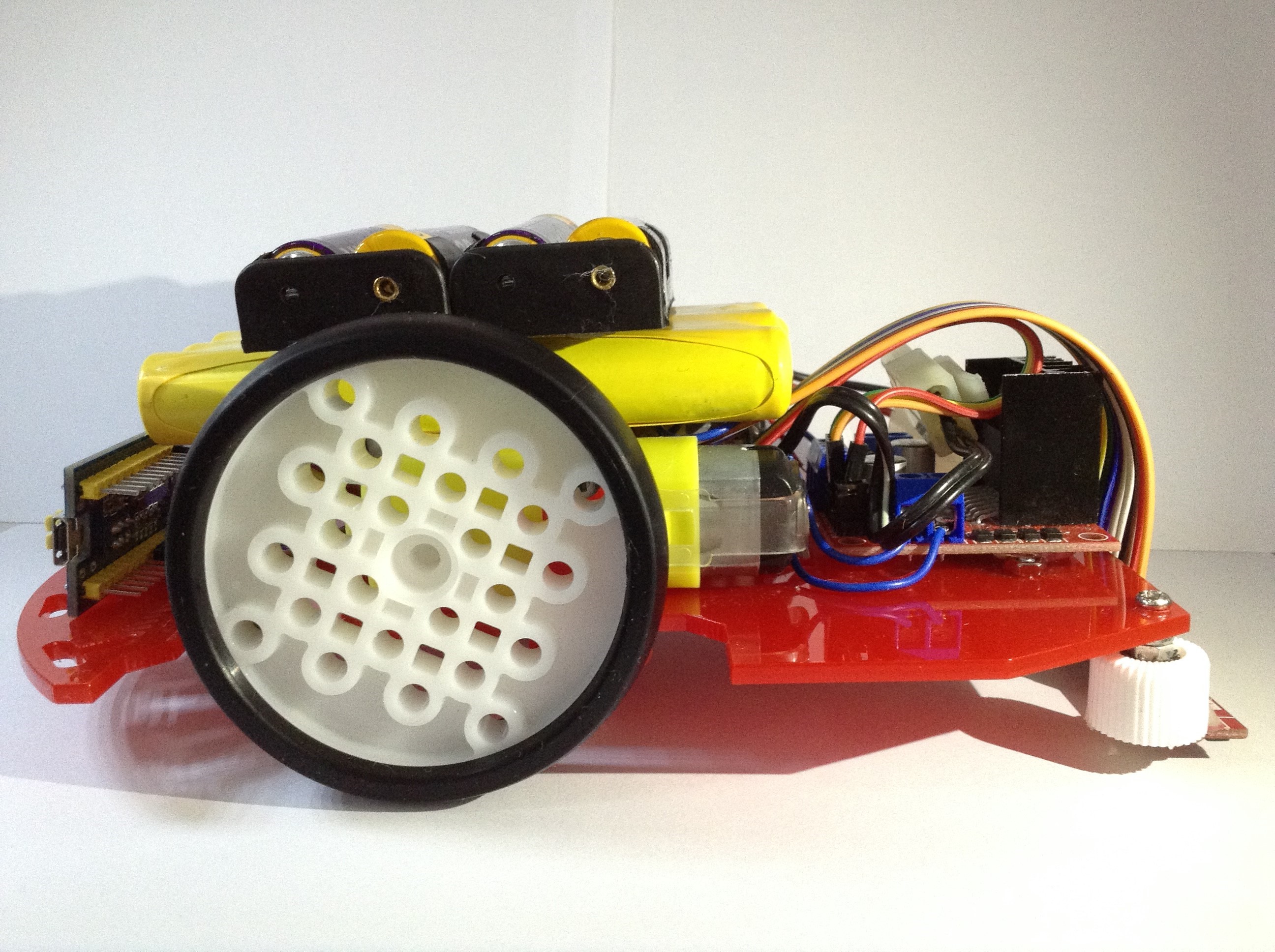}}%
\hfill
\subcaptionbox{Rear view}{\includegraphics[width=0.32\linewidth]{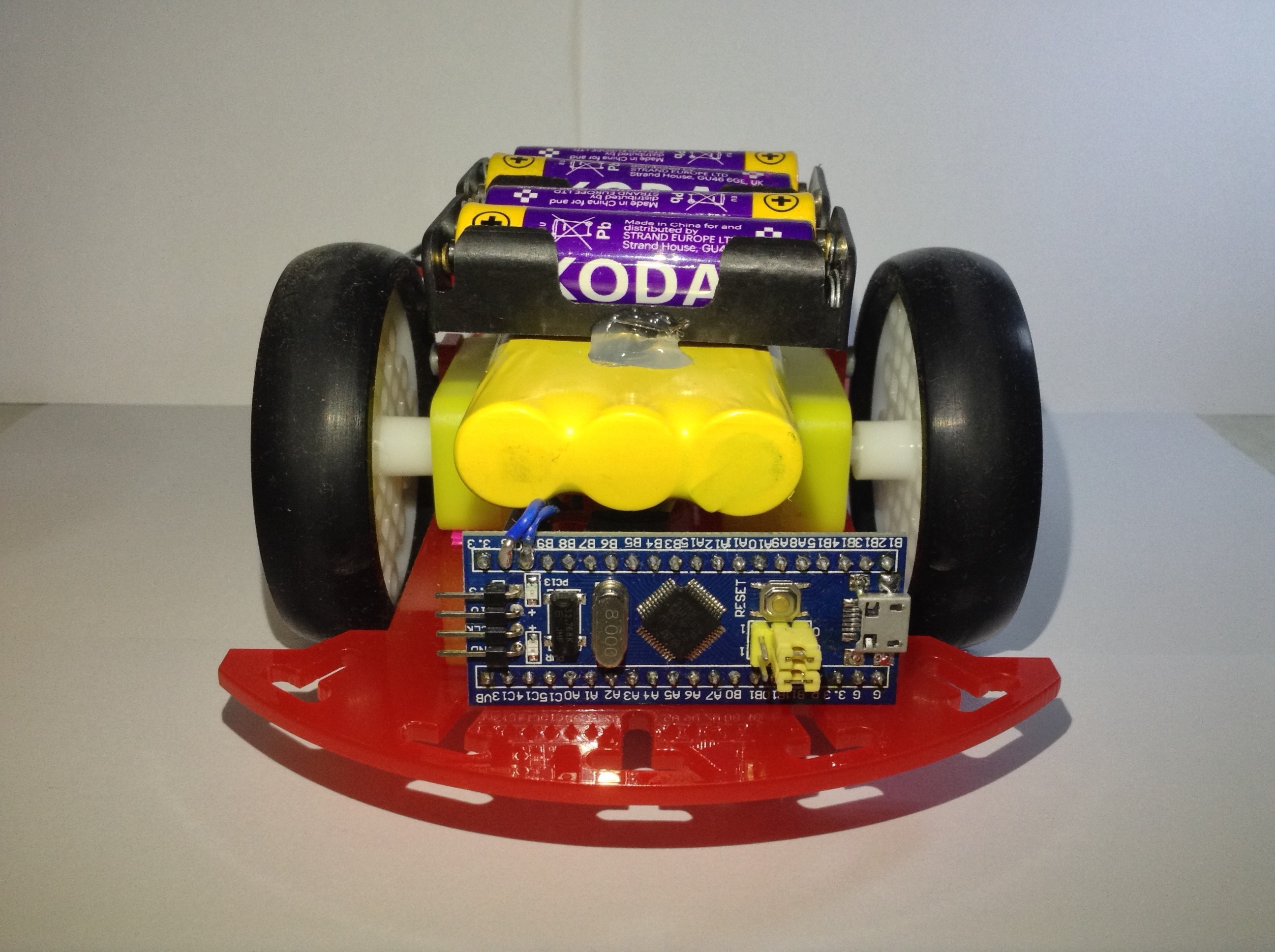}}%
\caption{Top, side and rear views of the line-tracking robots.}
\label{robots}
\end{figure*}

Fig.~\ref{robots} shows the two line-tracking robot used as a test platform throughout this research. Two low-cost 6V DC gear motors have been used. These motors spin at 250 RPM with 6 - 7 V DC input voltage. Since they have a gearbox, they generate enough torque required for faster responses. One of the advantages of these motors is that they are low-cost and a higher input voltage can be applied to reach a higher RPM value for short operation times. The DC motors have been mounted to the frame by using two plastic parts and two screws for each.

In line-following robots, wheels play an important role in the robot's handling of the path. High roadholding is needed to prevent the robot from slipping, especially in sharp turns. The wheels have been selected to minimize slipping and increase the robot's tracking capabilities of the path. Wheels with a diameter of 65 mm have been used to obtain high torque and minimize the effect of disturbances caused by the track. 

The frame enables the motors, sensors, motor driver used in the line-following robot to stay as a whole. The frame is made of a hard plastic with 174 mm length and 110 mm width. It is an off-the-shelf and low-cost component. The center of mass directly affects stability of the robot. By adjusting the location of the battery on the chassis, the center of mass can be manipulated since the heaviest component is battery. As the center of mass gets closer to the front of the robot, then slipping increases. On the other hand, as the center of mass gets closer to the rear, then the front side goes up and down in sharp turns. Hence, the center of mass has been centered relative to the center of the frame. Additionally, the distance between the ground and the chassis has been kept short which is another aspect affecting stability especially in sharp turns. In this way, it has also become easier to place the QTR-8 reflectance sensor closer to the ground.

\subsection{Electronic Components}

\subsubsection{Motor Driver}

To drive the two DC motors, a motor driver module having L298N dual full-bridge driver has been used. This motor driver module is supplied with 13 V DC voltage. It also has a 5 V DC voltage regulator which is used to supply STM32. There are 4 input pins that PWM signals are applied to so that the direction of rotation and the speed of the motors can be controlled. The output pins are dedicated for DC motors and can supply maximum 2 A current for each. The motor driver module has been mounted in front of the robot.

\subsubsection{Microcontroller}

To read sensor values and drive the DC motors accordingly, a microcontroller is needed. In this research STM32F103C8 microcontroller has been used. The system clock has been set to 72 MHz. APB2 timer clocks work at 72 MHz. Prescaler is 79 and counter period is 999 so that a 900 Hz PWM signal is generated. The duty cycle is manipulated adaptively by the PID controller. 4 pins of the microcontroller have been dedicated for PWM generation.

\subsubsection{Battery}

6 NiMH rechargeable batteries along with 4 AA alkaline batteries in series are used to power the robot. In this way, a total of 13 V DC input voltage is obtained. The positive terminal of the battery is connected to the 12 V input pin of the motor driver. It has been observed that at least 7 V input voltage is required to acquire 5 V output voltage from the internal regulator of the motor driver so that both STM32 and QTR-8 reflectance sensor work efficiently. The total value of current drawn by the system is approximately 1 Ampere. A great portion of this current is consumed by DC motors.

\begin{figure}
\centerline{\includegraphics[width=\linewidth]{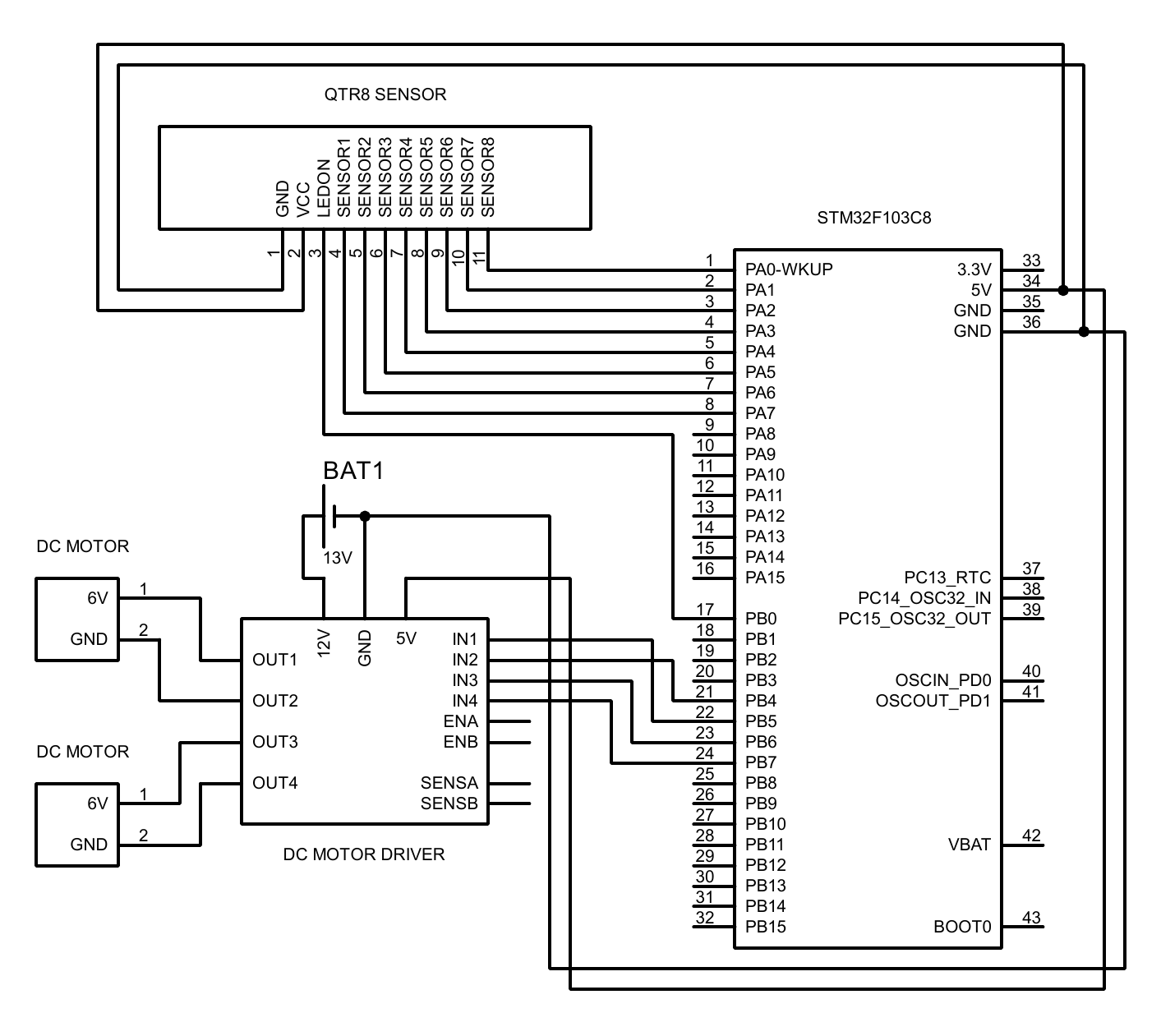}}
\caption{Circuit schematic of the line tracking robot.}
\label{schematic}
\end{figure}

\subsubsection{Sensor}

QTR-8 sensor is used for the robot to follow the line. There are eight reflectance sensors on this module. These sensors consist of infrared light emitters, MOSFETs switching in the presence of infrared light and capacitors. MOSFETs control the ground connection of capacitors. The 8 pins which correspond to a sensor are connected to STM32’s GPIO pins as shown in Fig.~\ref{schematic} for writing and reading operations. Before a reading operation, capacitors should be charged. For this, a logic 1 is written to the GPIO ports in output mode. If MOSFETs are exposed to infrared light, capacitors are connected to the ground, so they are discharged. Thus, after a few milliseconds the GPIO pins are switched to input mode. If the reading is logic 0 for a specific sensor, it means that there is a reflective surface. If the reading is logic 1, so there is a non-reflective, i.e. dark in colour, surface. This mechanism allows the microcontroller to understand where the line to be followed is. It should be noted that the distance between sensors and the track should not exceed 6 mm. Greater distance causes longer time taken for capacitors to discharge consequently reducing the sampling rate from sensors.

\section{System Characteristics}

\begin{figure}[t]
\centerline{\includegraphics[width=\linewidth]{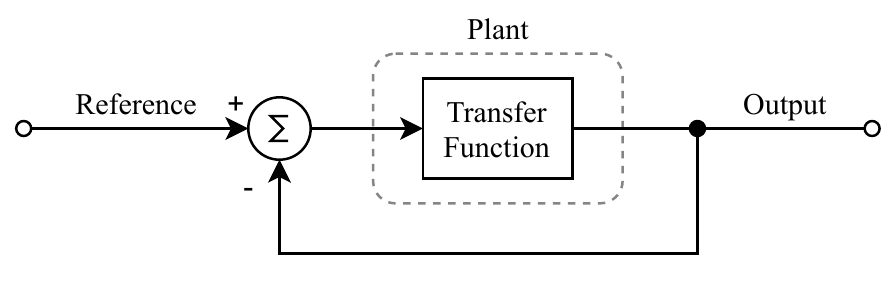}}
\caption{Block diagram of the unity feedback system used in system identification.}
\label{unity_feedback}
\end{figure}

Instead of using differential drive kinematics, a heuristic approach \cite{Balaji2015} has been carried out to identify the system characteristics. Before implementing a PID controller, a unity feedback system shown in Fig.~\ref{unity_feedback} was experimented on the robot. In paths where the line is straight, starting with non-zero error, the robot oscillates left and right by an equal amount. This implies that there is zero steady-state tracking error for a straight track. Consequently, it can be concluded that this system has at least one pole at origin. Then, the robot has been placed in the curved region of the track starting with non-zero position error. It has been observed that the system oscillates more towards the opposite side of the curvature. This implies that there is a steady-state error in following the curved path \cite{Balaji2015}. In this case, curvature can be thought of as a ramp or parabolic disturbance. This analysis shows that using an integral controller will add one more pole to the system enhancing its line-tracking capability in curved paths.

\section{Controller Design}

\begin{figure}[t]
\centerline{\includegraphics[width=7.5cm]{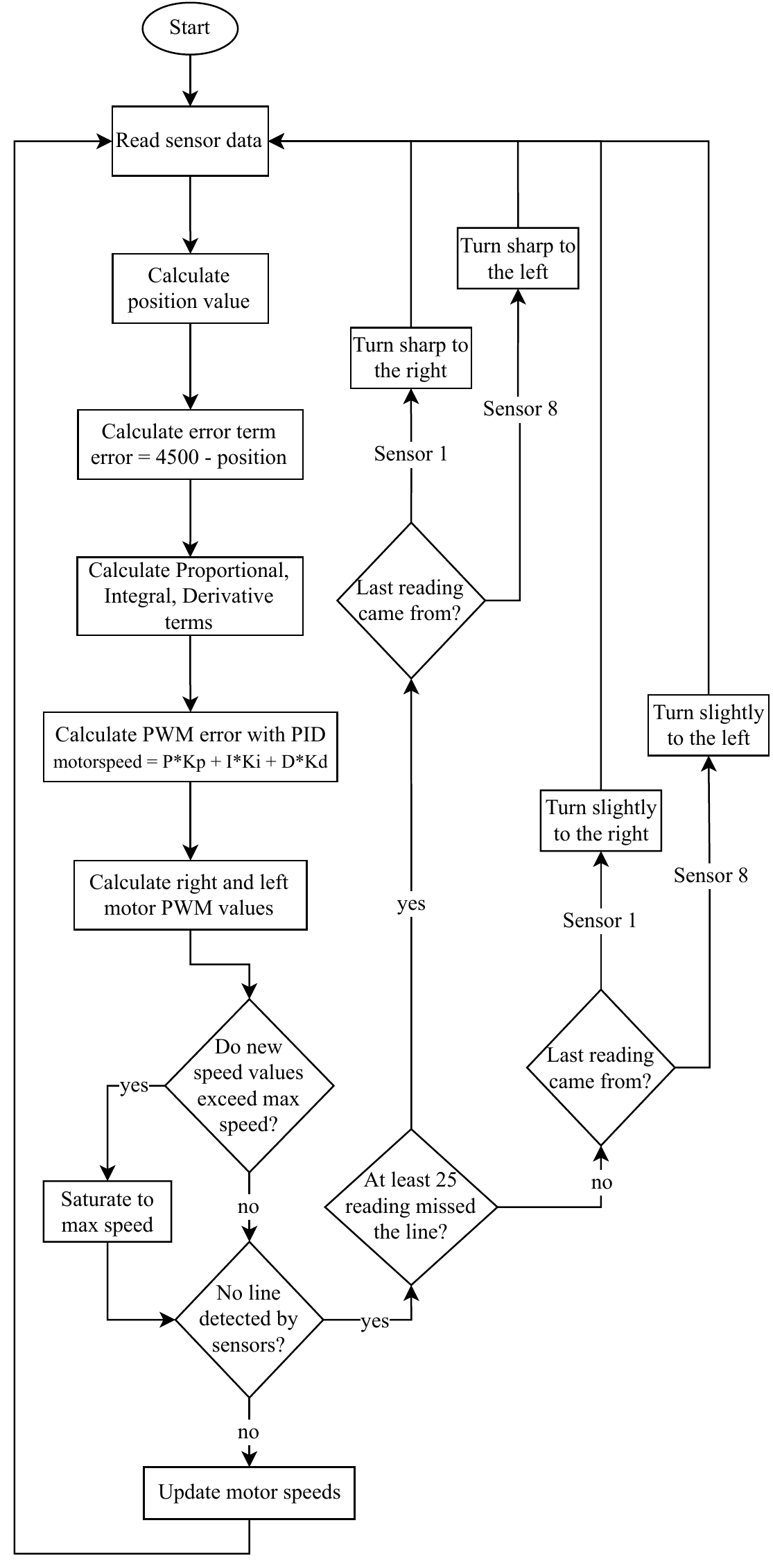}}
\caption{Flow chart of the line-tracking controller.}
\label{chart}
\end{figure}

The general structure of the controller algorithm is shown in Fig.~\ref{chart}. First, a reading is obtained from sensors. Based on this reading, a positional information is obtained so that the controller knows where the line is. Then, error is calculated and fed to the PID controller which manipulates the error term used for updating motor speeds. However, before feeding new PWM values, an open-loop control checks if sensors miss the line. Open-loop control is required to handle the corners in the test track. It also improves system performance in curved paths. 

\subsection{Calculation of the Positional Error}

Error is calculated based on sensor data. Each sensors’ reading corresponds to a position information. This position information is calculated by using \eqref{position}. The reference position value is 4500 indicating the middle of the sensor. Error is calculated by \eqref{error}. 

\begin{equation}
x = \frac{\sum_{n=1}^{8} 1000\cdot S_{n}}{\sum_{n=1}^{8} S_{n}} \label{position}
\end{equation}

\begin{equation}
e = 4500 - \frac{\sum_{n=1}^{8} 1000\cdot S_{n}}{\sum_{n=1}^{8} S_{n}} \label{error}
\end{equation}

If the error value is negative, this indicates that the robot is sliding to the right. If the error value is positive, the robot deviates from the line to the left. PID controller manipulates the speed of the motors. If the calculated error is negative, the speed of the motor on the left is reduced, the speed of the motor on the right is increased, and by this, the robot goes to the left side and eliminates the error. If the error value is positive, the speed of the motor on the left is increased and the speed of the motor on the right is decreased. Then the robot goes to the right side and the error is eliminated.

\subsection{Proportional Control}

Equation \eqref{p_control} is the mathematical expression for proportional control. It has been observed that when only P control is used, the robot responds faster to errors occurring during line tracking. But when the \emph{k\textsubscript{P}} gain goes above a certain value, oscillation has been found to be quite high. When the \emph{k\textsubscript{P}} gain is decreased, the oscillation decreases, but the robot response to errors has been found to be slower. For this reason, derivative control has been added to the system.

\begin{equation}
u(t) = k_{P}e(t) \label{p_control}
\end{equation}

\subsection{Integral Control}

When proportional control is used alone, it has been observed that a steady-state error occurs. The mathematical expression for the integral control is given by \eqref{i_control}. Integral term represents the summation of the past errors. Consequently, combining with \emph{k\textsubscript{I}} integral gain, integral  feedback minimizes steady-state tracking error by introducing a pole at origin \cite{Franklin1994}. 

\begin{equation}
u(t) = k_{I}\int_{t_{0}}^{t} e(\tau) \,d\tau \label{i_control}
\end{equation}

As investigated in the system identification stage, the unity feedback implementation without PID controller revealed that this system already has at least one pole at origin. However, the system is not able to follow the curved paths without steady-state error. To minimize the steady-state tracking error in curved paths, the integral term is added to the controller. However, it has been ineffective to sum all the past errors beginning from the start since the shape of the path is not always straight. To improve the response of the integral control for changes in the path’s curvature, only past five errors are considered \cite{Balaji2015}. So, the classical integral control has been modified as shown in \eqref{i_control_m}.

\begin{equation}
u(n) = k_{I}\times \sum_{n=0}^{4} e(n) \label{i_control_m}
\end{equation}

\subsection{Derivative Control}

Equation \eqref{d_control} is the mathematical expression for the classical derivative control. Derivative control has been added to eliminate oscillation and overshooting caused by proportional and integral terms. 

\begin{equation}
u(t) = k_{D}\times\frac{de(t)}{dt} \label{d_control}
\end{equation}

D control prevents overshooting in the system and improves the system's transient response. However, it should be noted that the D control is sensitive to noise \cite{Franklin1994}. The reason why D control is sensitive to noise is that it acts like a high pass filter amplifying instant changes in error. The effect of noise has not been seen much during the development of the robot. So, the classical derivative has been modified as in \eqref{d_control_m} so the calculation is done by finding the difference between the current and previous error.

\begin{equation}
u(n) = k_{D}\times (e(n) - e(n-1)) \label{d_control_m}
\end{equation}

\subsection{Modified PID Controller}

Equation \eqref{pid_control} expresses the classical PID control. With the aforementioned modifications, the overall mathematical expression for the PID controller is given by \eqref{pid_control_m}. In this controller, proportional term is used to minimize positional error, integral term is used to minimize steady-state error occurring in curved paths and derivative term is used to minimize oscillatory behavior and overshooting.

\begin{equation}
u(n) = k_{P}e(t) + k_{I}\int_{t_{0}}^{t} e(\tau) \,d\tau + k_{D}\times\frac{de(t)}{dt} \label{pid_control}
\end{equation}

\begin{equation}
u(n) = k_{P}e(n) + k_{I} \sum_{n=0}^{4} e(n) + k_{D} (e(n) - e(n-1)) \label{pid_control_m}
\end{equation}

\subsection{Open-loop Controller}

The PID controller shows good performance in straight lines as well as curved paths. However, for handling sharp turns, it shows a poor performance, that is the robot goes off the track. Especially, in 45 degree turns, the robot was turning opposite to the path when only the PID controller was used. A thorough analysis has been conducted to understand why the PID controller fails in sharp turns. The analysis has revealed that in fact, the PID controller initially responds to the change in the path correctly for a short time. However, when all sensors miss the line, last errors come from the opposite side of the path due to the initial response of the PID controller causing the robot to miss the track. Therefore, an open-loop control mechanism has been developed for handling sharp turns. In this control, when all the sensors miss the line, only right and left end sensors, i.e. sensor 1 and sensor 8 respectively, are considered to understand which direction the robot should turn. If the last error reading comes from sensor 1, the robot should turn right, if it comes from sensor 8, the robot should turn left by rotating the two wheels in opposite directions. The amount of speed at which each motor will rotate is determined by another control mechanism developed to handle both curved paths and corners as fast as possible. If the sensors miss the line for at least 25 reading which is the case when handling corners, then a sharp turn is performed, i.e. inner wheel rotates backwards at 53\% of the full speed and outer wheel rotates forward at 70\% of the full speed. This enables the robot to quickly come back to the line in 90 and 45 degree turns.  If the sensors miss the line for a few times, i.e. not more than 25, which is the case when handling curved paths, then the inner motor rotates backward at 20\% of the full speed and outer wheel rotates forward at full speed. This causes the robot to turn slightly and go faster where the path is curved. Even though integral control shows a good performance in handling curvatures, this open-loop control improves the performance when handling slight turns.

\section{Analysis}

\begin{figure}[t]
\centerline{\includegraphics[width=7.5cm]{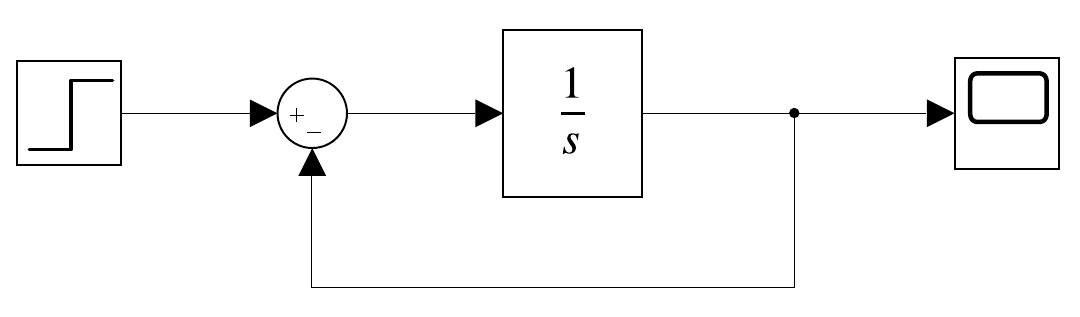}}
\caption{A Type 1 unity feedback system.}
\label{unity_diagram}
\end{figure}

\begin{figure}[t]
    \begin{subfigure}{\linewidth}
    \centerline{\includegraphics[width=8cm]{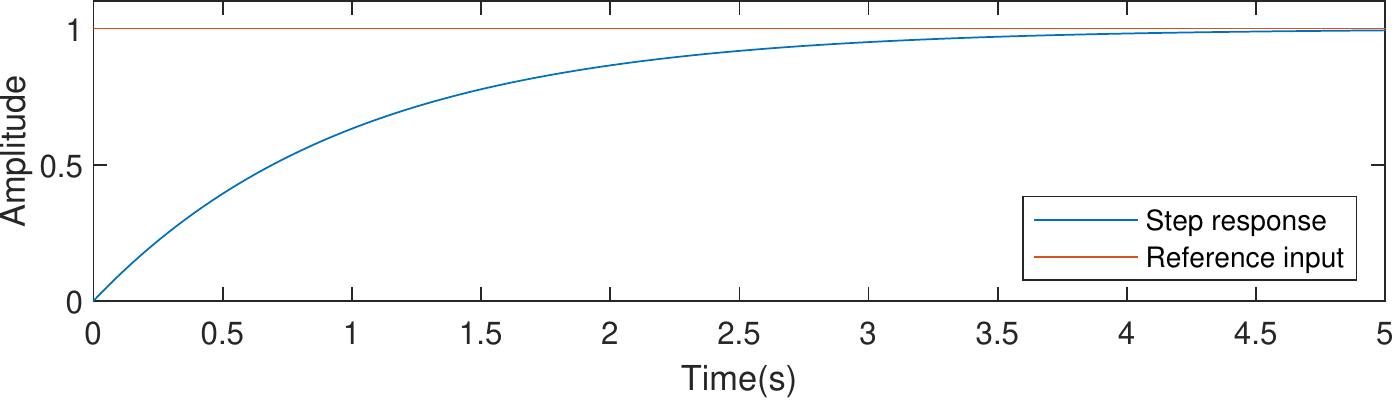}}
    \caption{Step response of the Type 1 unity feedback system.}
    \label{unity_step}
    \end{subfigure}\par\medskip
    
    \begin{subfigure}{\linewidth}
    \centerline{\includegraphics[width=8cm]{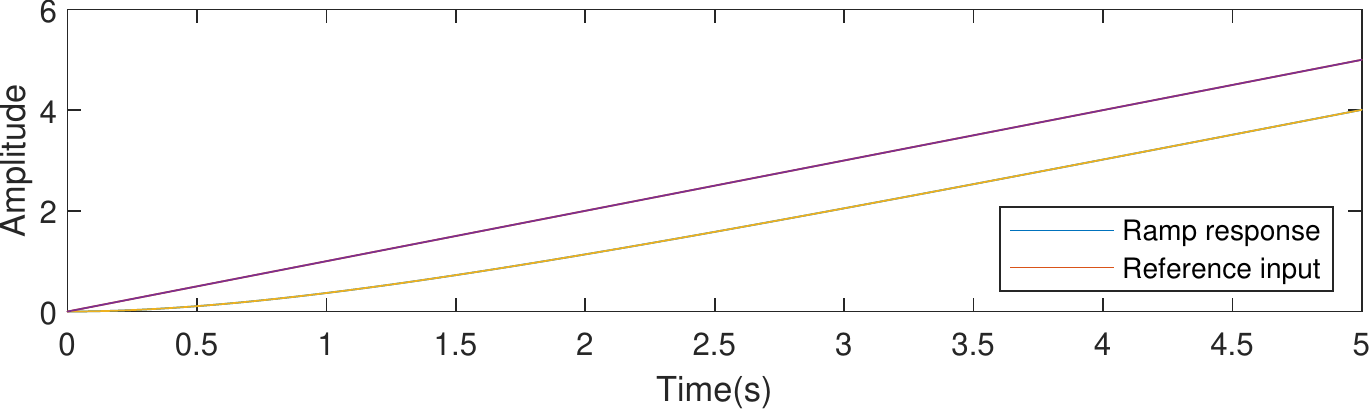}}
    \caption{Ramp response of the Type 1 unity feedback system.}
    \label{unity_ramp}
    \end{subfigure}\par\medskip
    
    \begin{subfigure}{\linewidth}
    \centerline{\includegraphics[width=8cm]{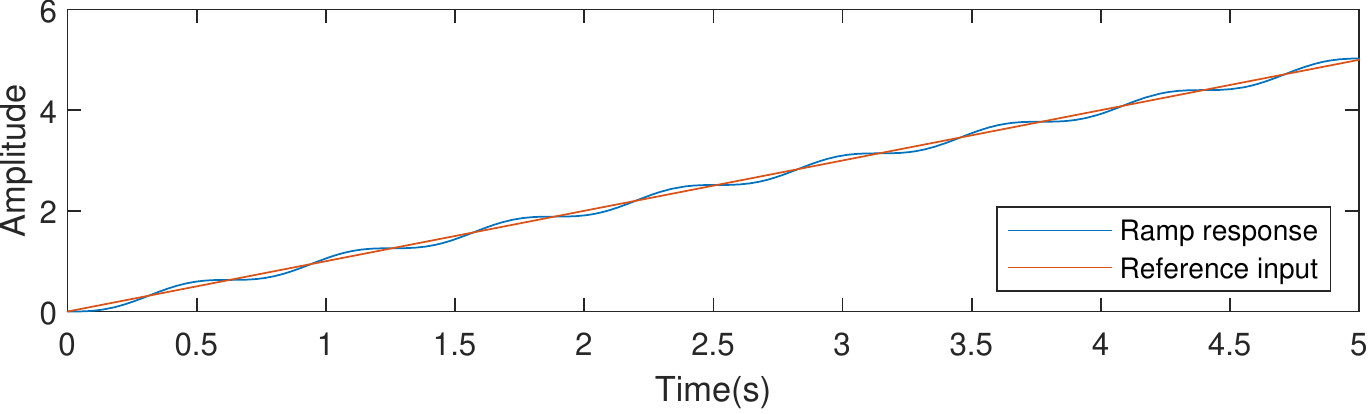}}
    \caption{Ramp response when integral term is added.}
    \label{int_ramp}
    \end{subfigure}
    \caption{Analyzing steady-state tracking error for different system configurations.}
\end{figure}

In this research, a heuristic approach \cite{Balaji2015} has been followed for system identification. It has been found that this system has a pole at origin. Furthermore, straight lines can be thought as step disturbance and curved paths can be thought as ramp disturbance. Step and ramp response of an arbitrary Type 1 transfer function has been simulated as shown in Fig.~\ref{unity_diagram} and analyzed to get a better understanding of the system under consideration. Fig.~\ref{unity_step} shows the step response of an arbitrary Type 1 unity feedback system. As expected, there is zero steady-state error for a step reference. However, Type 1 systems cannot track ramp references with zero steady-state error. Fig.~\ref{unity_ramp} shows the ramp response of the same Type 1 unity feedback system. Therefore, integral control is added to the system to obtain zero-steady state error when tracking ramp references. Fig.~\ref{int_ramp} shows the ramp response of the Type 1 system with integral control added.

\section{Results}

\begin{figure}[b]
\centerline{\includegraphics[width=7.5cm]{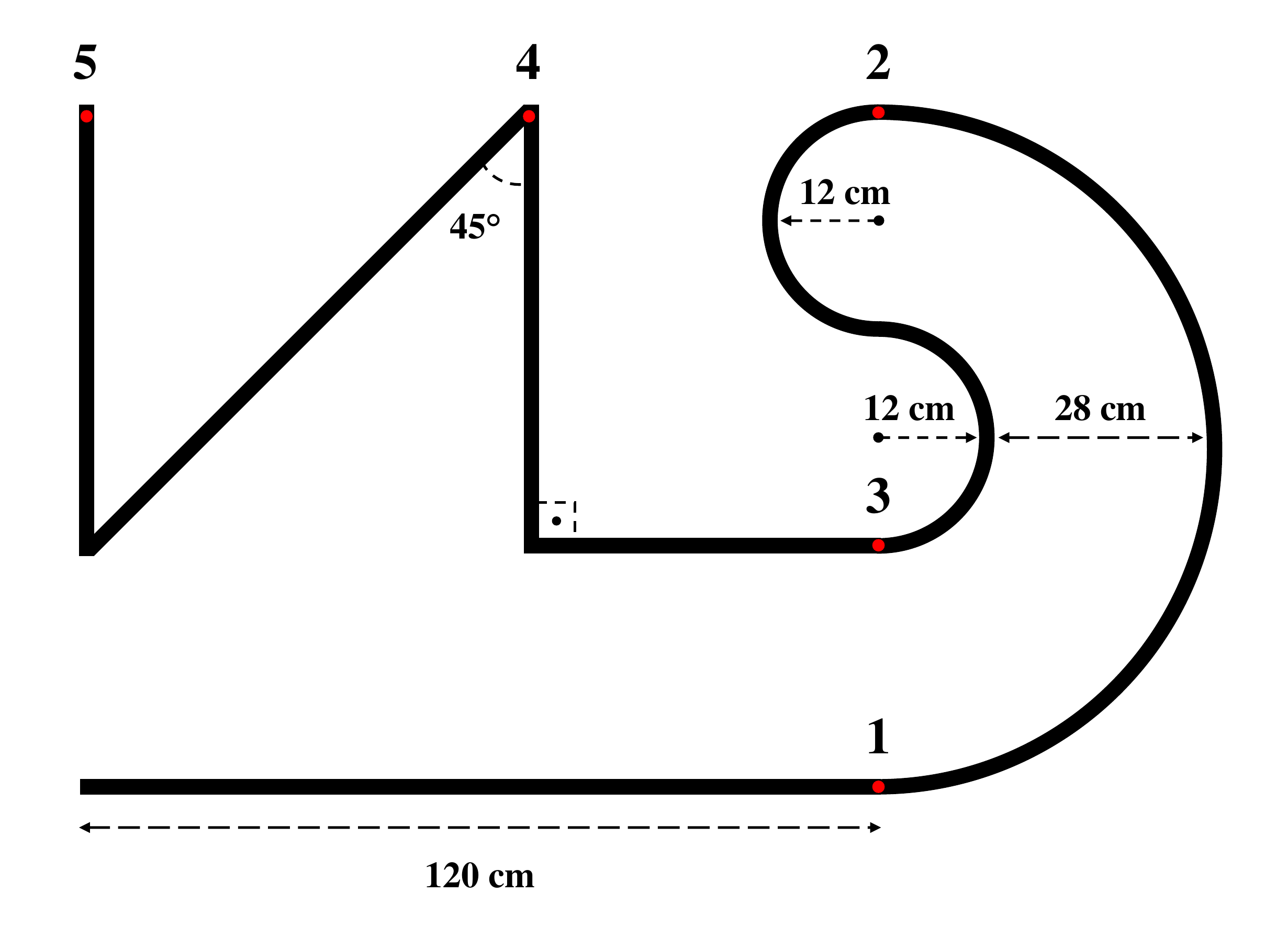}}
\caption{A slightly modified version of the test track in \cite{Balaji2015}.}
\label{track}
\end{figure}

Fig.~\ref{track} shows the test track consisting of straight lines, curvatures, and corners with 90 and 45 degree turns. The track is made of wooden plates having a light color. A black electrical tape has been used to create the path so that reflectance sensors can work efficiently.

Before implementing the open-loop controls, with 10.2 V battery voltage, the robot could not handle 45 degree turns. After implementing the open-loop control in which when all sensors miss the line, only leftmost and rightmost sensors are considered, the robot completed the track in 9.2 seconds. To further reduce the completion time, the battery voltage was increased up to 13 V. However, the robot started to fail to follow the line in sharp turns. For handling sharp turns and improving system performance in curved paths, the open-loop control has been modified. Finally, with 13 V battery voltage and 92\% base PWM duty cycle, the robot successfully completed the track in 7.8 seconds.

\section{Discussion}

Considering the limitations of the DC motors and wheels, it has been challenging to reduce the completion time below 7.5 seconds. It has been observed that dust and dirt on the test track and wheels negatively affect completion time and stability. When the track is dirty, it has been sometimes observed that the robot slipped off the track and could not complete it. Another limitation is that 6 V DC motors are slow. To address this issue, a higher input voltage has been applied to the motor driver, i.e. 13 V. Yet another difficulty was handling 45 degree turns. Initially, an integral control based approach has been developed in which only a few past errors were being considered after all sensors miss the line. However, this approach suffered from the initial response of the PID controller to the 45 degree turn. So, past errors were also coming from the opposite side of the path. To eliminate this problem, a basic method has been developed in which only right and left end sensors are considered to decide which direction to turn after all sensors miss the line. Another limitation is posed by the distance between each sensor on the QTR-8. The discrete structure of the sensor array causes the robot to oscillate while following the line. If another module having sensors placed more densely is used, it is expected that the amount of oscillation is reduced since the sudden changes in error is mitigated. Yet another limitation is that batteries are contributing to a great portion of the robot’s weight. Also they cause the robot skid off the path in sharp turns.

\section{Conclusion}

In this paper, a PID controller scheme along with additional open-loop controls has been developed for a line tracking robot consisting of low-cost, off-the-shelf components such as STM32F103C8, QTR-8RC reflectance sensor, L298N motor driver. During the construction of the robot, it has been ensured that sensors are close to the ground, i.e. 3.5 mm, and the total weight is approximately at the center of the chassis and close to the ground. Before designing a controller, the system characteristics have been identified with a simple unity feedback implementation. Then, a PID controller has been implemented. However, to improve system stability and robustness, a number of changes have been made. Since the PID controller was ineffective in handling sharp turns, an open-loop control mechanism has been developed. To minimize overshooting and oscillatory behavior, $ k_{p} $, $k_{I}$ and $k_{D}$ gains have been optimized through experiments. One of the main limitations affecting stability of the system is the separation between each sensor in the QTR-8 module. This separation causes the error term to change suddenly consequently the robot shows oscillatory behaviour.

\bibliographystyle{IEEEtran}
\bibliography{article}
\end{document}